\documentclass[conference]{IEEEtran}
\IEEEoverridecommandlockouts
    
\usepackage{cite}
\usepackage{amsmath,amssymb,amsfonts}
\usepackage{algorithmic}
\usepackage{graphicx}
\usepackage{textcomp}
\usepackage{xcolor}
\usepackage{xspace}
\usepackage{multirow}
\usepackage{algorithm}
\usepackage{algorithmic}
\usepackage{subfig}
\usepackage{soul}
\usepackage{makecell}
\usepackage{balance}
\def\BibTeX{{\rm B\kern-.05em{\sc i\kern-.025em b}\kern-.08em
    T\kern-.1667em\lower.7ex\hbox{E}\kern-.125emX}}

\newtheorem{definition}{Definition}
\newtheorem{example}{Example}

\newtheorem{proof}{Proof}

\newtheorem{theorem}{Theorem}

\newcommand{\eg}{\textit{e.g.}\xspace}
\newcommand{\ie}{\textit{i.e.}\xspace}

\newcommand\figref[1]{Fig.~\ref{#1}}
\newcommand\algref[1]{Algorithm~\ref{#1}}
\newcommand\tabref[1]{Tab.~\ref{#1}}
\newcommand\secref[1]{Sec.~\ref{#1}}

\newcommand{\fakeparagraph}[1]{\vspace{1mm}\noindent\textbf{#1.}}
\newcommand{\sysname}{FOCUS\xspace}

\ifodd 0

\newcommand{\blue}[1]{{\color{blue}{#1}}}
\else

\newcommand{\blue}[1]{{{#1}}}
\fi

\begin{document}



\title{Accurate and Efficient Multivariate Time Series Forecasting via Offline Clustering}

\author{
	\IEEEauthorblockN{
		Yiming Niu\IEEEauthorrefmark{1}, 
		Jinliang Deng\IEEEauthorrefmark{2}\IEEEauthorrefmark{3}, 
		Lulu Zhang\IEEEauthorrefmark{1}, 
		Zimu Zhou\IEEEauthorrefmark{4} 
		Yongxin Tong\IEEEauthorrefmark{1}} 
	\IEEEauthorblockA{\IEEEauthorrefmark{1}State Key Laboratory of Complex \& Critical Software Environment,\\
Beijing Advanced Innovation Center for Future Blockchain and Privacy Computing, Beihang University, Beijing, China \\}
\IEEEauthorblockA{\IEEEauthorrefmark{2}HKGAI, Hong Kong University of Science and Technology, Hong Kong SAR, China}
\IEEEauthorblockA{\IEEEauthorrefmark{3}Research Institute of Trustworthy Autonomous Systems, Southern University of Science and Technology, Shenzhen, China}
	\IEEEauthorblockA{\IEEEauthorrefmark{4}Department of Data Science, City University of Hong Kong, Hong Kong SAR, China} 
    \{yimingniu@, zhangluluzll@, yxtong@\}buaa.edu.cn\\
    dengjinliang@ust.hk, zimuzhou@cityu.edu.hk
}

\maketitle

\begin{abstract}

Accurate and efficient multivariate time series (MTS) forecasting is essential for applications such as traffic management and weather prediction, which depend on capturing long-range temporal dependencies and interactions between entities. Existing methods, particularly those based on Transformer architectures, compute pairwise dependencies across all time steps, leading to a computational complexity that scales quadratically with the length of the input.
To overcome these challenges, we introduce the Forecaster with Offline Clustering Using Segments (\sysname), a novel approach to MTS forecasting that simplifies long-range dependency modeling through the use of prototypes extracted via offline clustering. These prototypes encapsulate high-level events in the real-world system underlying the data, summarizing the key characteristics of similar time segments.
In the online phase, \sysname dynamically adapts these patterns to the current input and captures dependencies between the input segment and high-level events, enabling both accurate and efficient forecasting. By identifying prototypes during the offline clustering phase, \sysname reduces the computational complexity of modeling long-range dependencies in the online phase to linear scaling. Extensive experiments across diverse benchmarks demonstrate that \sysname achieves state-of-the-art accuracy while significantly reducing computational costs.

\end{abstract}

\begin{IEEEkeywords}
multivariate time series, spatiotemporal data mining, forecasting
\end{IEEEkeywords}

\section{Introduction}
\label{sec:intro}

Accurate and efficient multivariate time series (MTS) forecasting is of great importance in various real-world applications~\cite{ariyo2014arima, lai2018energy, cheng2021windturbine, cirstea2021enhancenet, wu2021autocts, yan2021flashp, wu2023autocts+, yao2023simplets, shao2024exploring}. 
MTS data, characterized by its complex structure encompassing both temporal and entity dimensions~\cite{wu2021autocts, deng2022multi, lai2023lightcts, ye2022learning, deng2024disentangling, deng2024parsimony}, is crucial for enabling precise predictions across different application scenarios.
For example, in traffic flow forecasting, accurate predictions can optimize resource allocation~\cite{cirstea2022towardstraffic, cirstea2021enhancenet, wu2021autocts}, while in weather forecasting, efficient predictions can provide timely warnings to the public~\cite{wang2019weatherforecasting, he2022climateforecasting}.

A key challenge in achieving accurate MTS forecasting lies in modeling long-range dependencies~\cite{zhou2021informer, nie2023patchtst, guo2023selfsup, li2023towards}. 
These dependencies capture the deep relationships across long time periods (\eg seasonal variations of temperature) and multiple entities (\eg traffic flow relationships between intersections) within the data~\cite{zha2024scaling, jiang24sagdfn, he2023oneshotstl, cheng2023weakly}.
Modeling these long-range dependencies involves two primary steps: First, identifying potential \textit{events} within the data, such as peaks in traffic or fluctuations in climate. 
Each event corresponds to a cluster of similar time segments, where each cluster shares a representative segment pattern, referred to as a \textit{prototype}. 
Next, these prototypes and their corresponding events are analyzed to further explore their temporal relationships and interactions between entities. This detailed modeling helps capture the long-range dependencies, thereby enabling accurate time series forecasting.

Modeling long-range dependencies in MTS poses a critical challenge.
Canonical neural architectures, such as CNNs and RNNs, require increasingly more computational steps (hops) to connect distant time points as the time lag grows, making them ineffective for capturing long-range dependencies.
In contrast, Transformer architectures~\cite{vaswani2017attention, wu2021autoformer, zhou2022fedformer}, with their ability to directly model global correlations across sequences of arbitrary lengths, offer a promising solution to this problem.
PatchTST~\cite{nie2023patchtst} models the long-range dependencies between time segments, significantly improving forecasting accuracy compared with previous solutions. 
Although this approach implicitly identifies representative segment patterns during the learning process and models their dependencies, it does not utilize these patterns effectively. 
Instead, it still treats all segments as independent units, leaving the model confined to all-pairs dependency modeling framework, which results in $\mathcal{O}(L^2)$ computational complexity.

To enhance computational efficiency, Informer~\cite{zhou2021informer} introduces the ProbSparse self-attention mechanism, which performs sparse sampling on the attention matrix and computes attention only for critical key-value pairs, reducing complexity to $\mathcal{O}(L \log L)$. However, this sampling mechanism may discard critical information, resulting in performance degradation when capturing intricate patterns.  
Crossformer~\cite{zhang2023crossformer} leverages low-rank properties to compress input sequences and decompose the original large attention matrix into the product of two smaller matrices, reducing computational complexity to $\mathcal{O}(2kN)$, where $N$ represents the number of entities. However, it heavily relies on dimensionality reduction.
Although these methods reduce computational costs through sampling and dimensionality reduction, they fail to fundamentally eliminate the reliance on all-pairs dependency modeling on segments. 
As a result, the efficiency and accuracy trade-off remains, offering limited practical value in addressing the core issue.

\begin{figure}[t]
    \centering
    \includegraphics[width=\linewidth]{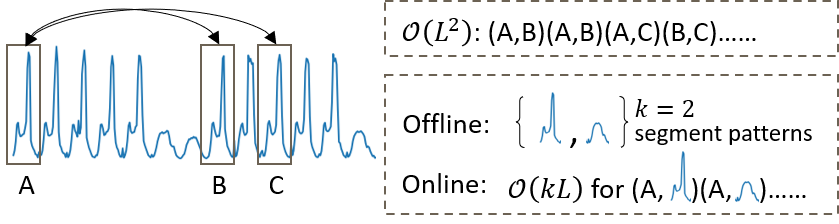}
    \caption{Modeling long-range dependency on Traffic dataset~\cite{wu2021autoformer} with representative segment patterns (\ie prototypes) discovered offline.}
    \label{fig:core_idea}
\end{figure}

A promising solution to address the complexity of long-range dependency modeling is to shift the identification of prototypes to an offline phase. 
Given specific applications, these prototypes are relatively universal and can be effective across various instances.
Therefore, we can extract these prototypes offline from the dataset and apply them to different instances in the online phase, as shown in \figref{fig:core_idea}. 

\fakeparagraph{Our Solution}
We propose the \textbf{F}orecaster with \textbf{O}ffline \textbf{C}lustering \textbf{U}sing \textbf{S}egments (\sysname), a model that leverages prototypes from time series data offline. This approach simplifies modeling long-range dependencies in forecasting. The FOCUS operates in two distinct phases:

In the \textit{offline clustering}, the focus is on extracting prototypes from the dataset. The process begins by dividing the entire training dataset's time series data into smaller segments. 
These segments are then clustered based on their similarities, with each cluster represented by a prototype that summarizes the overall pattern of the cluster. 
To ensure that these prototypes accurately reflect the features of their respective clusters, an iterative optimization process is employed. 
This process not only evaluates the distance-based similarity between the prototype and the segments but also considers the relational dependencies within the cluster. 
But there may be discrepancies between the prototypes and specific instances, suggesting a need to adapt the prototypes based on the instance.

In the \textit{online adaptation}, the focus shifts to how the extracted prototypes can be adapted to the online input instances to achieve more accurate predictions. The online input is segmented in a manner similar to the offline phase, with each segment dynamically assigned to its closest prototype. This adaptation allows the prototypes to be tailored to current inputs. A deep learning model then establishes associations between the input segments and the adapted prototypes, effectively modeling the long-range dependencies. Based on these associations, the model performs precise forecasting.

\fakeparagraph{Contributions}
Our contributions are summarized as follows: 
\begin{itemize} 
\item We propose a novel forecasting model, \sysname, that combines an offline clustering phase to identify representative segment patterns and an online adaptation phase to dynamically refine these patterns based on input data, ensuring accurate and efficient forecasting.
\item We design an efficient mechanism to capture long-range dependencies with linear complexity by leveraging the correlations between a fixed set of representative segment patterns (\ie prototypes) and the input sequence. This enables the model to handle long and high-dimensional sequences efficiently without compromising accuracy.
\item Comprehensive experiments on diverse datasets demonstrate that \sysname outperforms state-of-the-art methods in terms of computational efficiency, while ranking top-1 accuracy among the 8 models for comparison on 26 out of the 28 settings.
\end{itemize}

The remainder of this paper is organized as follows:  
In \secref{sec:motivation}, we discuss the motivation for addressing long-range dependency challenges.  
\secref{sec:overview} provides an overview of our method, followed by details on offline clustering in \secref{sec:offline} and online adaptation in \secref{sec:online}.  
We describe the dual-branch architecture for efficient feature fusion in \secref{sec:dual-branch}.  
Finally, \secref{sec:experiments} presents experimental results demonstrating the efficiency and accuracy of \sysname.

\section{Problem Statement}
\label{sec:preliminaries}

\begin{table}[h]
\centering
\caption{Frequently used notations.}
\label{tab:frequently_used_notations}
\begin{tabular}{c l}
\hline
\textbf{Notation} & \textbf{Description} \\ \hline
$d$ & the intrinsic dimension of features within one token \\ 
$k$ & the number of prototypes for the offline and online process \\
$p$ & Number of time steps within one segment \\
\blue{$N$} & The total entities contains in the dataset \\
$T$ & The total time steps contains in the dataset \\
$L$ & Number of historical time steps (lookback) \\ 
\blue{$L_f$} & Number of future time steps (horizon) \\ 
\blue{$l$} & Number of segments \\ 
$\mathcal{D}$ & The time series dataset \\
$\mathcal{X}$ & Historical multivariate time series data \\ 
$\mathcal{X}_{e, t}$ & Value of variable $c$ at time step $t$ \\ 
$\mathcal{Y}$ & Actual future multivariate time series data \\ 
$\hat{\mathcal{Y}}$ & Predicted future multivariate time series data \\ 
$\mathcal{P}$ & A matrix of time series segments, each of length \( l \).\\ 
$\mathcal{P}_{e, i}$ & The $i$-th segment of entity $e$.\\ 
$\mathcal{M}$ & Model used for forecasting \\ 
$\mathcal{H}$ & The hidden state of multiple entities and multiple tokens\\
\hline
\end{tabular}
\label{table:notations}
\end{table}

\begin{figure*}[t]
    \centering
    \includegraphics[width=0.8\linewidth]{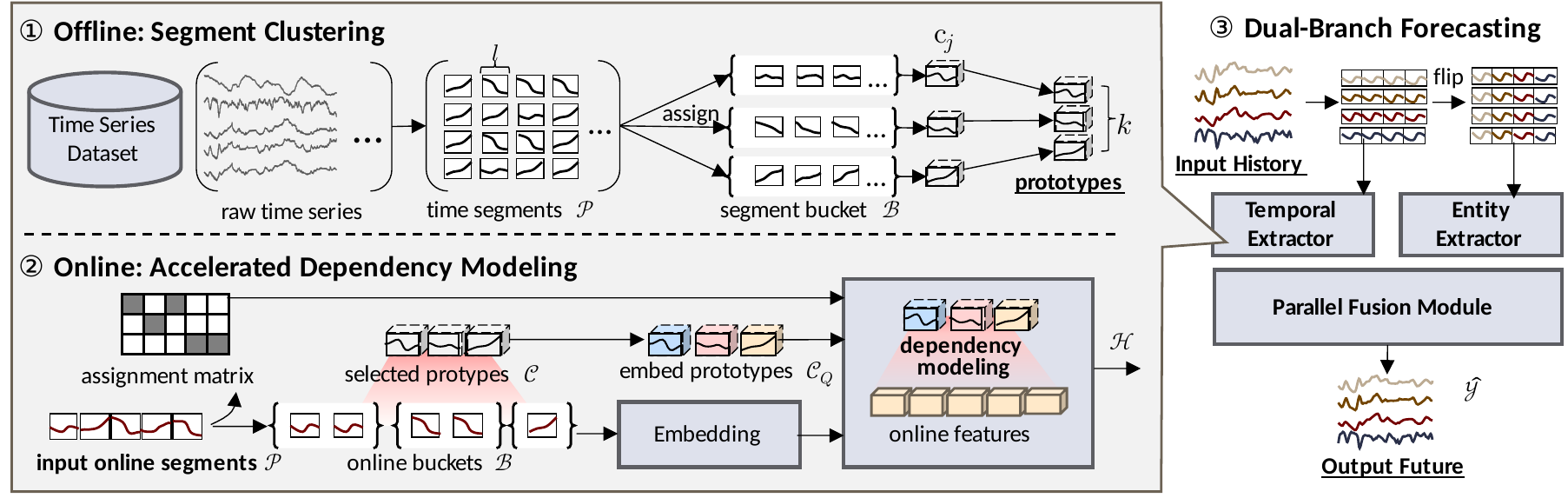}
    \caption{Overview of the proposed \sysname.}
    \label{fig:overview}
\end{figure*}

We first formally define the time series data and then define the multivariate time series forecasting problem. \tabref{tab:frequently_used_notations} shows the frequently used notations.

\definition{\textit{(Time Series)}}{
A time series represents a collection of attributes over time for a single variable. Formally, it is defined as:
\begin{equation}
\mathcal{D}_e = \{ x_{e,1}, x_{e,2}, \ldots, x_{e,T} \},
\end{equation}
where \(\mathcal{D}_e\) denotes the data of variable \(e\) over \(T\) time steps, and \(x_{e,t}\) is the value of variable \(e\) at time step \(t\).
}

\definition{\textit{(Multivariate Time Series)}}{
A multivariate time series is a collection of time series for multiple variables or entities. For \(N\) variables or entities, it is defined as:
\begin{equation}
\mathcal{D} = \{ \mathcal{D}_1, \mathcal{D}_2, \ldots, \mathcal{D}_D \},
\end{equation}
\blue{where each $\mathcal{D}_e = \{ x_{e,1}, x_{e,2}, \ldots, x_{e,T} \}$ represents the sequence of values for variable $e$ over $T$ time steps. Additionally, for any two variables $e_1$ and $e_2$ ($1\leq e_1\neq e_2\leq N$), the time index $t$ is consistent across all variables, meaning that the time step corresponding to $x_{e_1,t}$ is the same as the time step corresponding to $x_{e_2,t}$ for $t = 1,2,\ldots,T$. Relationships or dependencies may exist between the variables.}
}

\definition{\textit{(Multivariate Time Series Forecasting)}}{
The goal of multivariate time series forecasting is to predict future segments of the series based on observed historical segments. 
Given the historical data \(\mathcal{X}\), which is a subset of the multivariate time series dataset \(\mathcal{D}\) over a specific time period of length \(L\), it is defined as:
\begin{equation}
\mathcal{X} = \{ \mathbf{x}_1, \mathbf{x}_2, \ldots, \mathbf{x}_L \} \in \mathbb{R}^{D \times L},
\end{equation}
where \(\mathbf{x}_t \in \mathbb{R}^N\) represents the values of \(D\) variables at time step \(t\). The goal is to predict the future time series \(\mathcal{Y}\):
\begin{equation}
\mathcal{Y} = \{ \mathbf{y}_1, \mathbf{y}_2, \ldots, \mathbf{y}_{L_f} \} \in \mathbb{R}^{N \times L_f},
\end{equation}
where \(L_f\) is the number of future time steps, and \(\mathbf{y}_t \in \mathbb{R}^N\) represents the values of \(N\) variables at future time step \(t\). The forecasting task aims to find a model \(\mathcal{M}\) such that:
\begin{equation}
\hat{\mathcal{Y}} = \mathcal{M}(\mathcal{X}),
\end{equation}
where \(\hat{\mathcal{Y}}\) denotes the predicted future time series.
}

\section{Motivations}
\label{sec:motivation}

Modeling dependencies among all segments in multivariate time series theoretically captures all long-range dependencies, but its high computational complexity makes it impractical.  
This approach has quadratic computational complexity, resulting in excessive resource and time consumption when processing long and high-dimensional data. 
Thus, alternative methods are needed to reduce computational complexity.

To address this, we can explore representative segment patterns within the data as a more efficient solution.
Representative segment patterns are representations of high-level system events, effectively capturing temporal and spatial repetition.
These patterns exhibit stable recurrence over time and space, serving as key features of high-level system events.

\begin{figure}[t]
    \centering
    \includegraphics[width=0.8\linewidth]{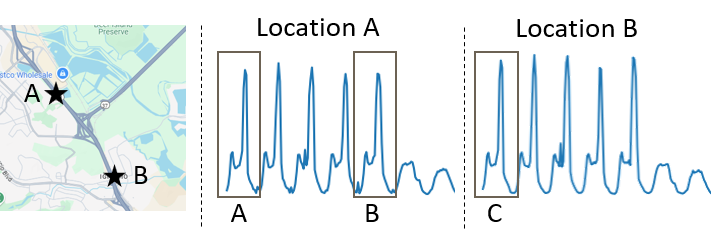}
    \caption{Example from Traffic dataset~\cite{wu2021autoformer} of a 7-day period.}
    \label{fig:motivation}
\end{figure}

\example{ 
\textit{ 
Consider a 7-day traffic flow time series as shown in \figref{fig:motivation}, which can be divided into fixed-length intervals (e.g., hourly windows).  
These intervals typically exhibit simple, predictable patterns, such as increased congestion during rush hours and decreased flow at night.  
While these patterns may vary slightly depending on specific conditions, their recurrence remains stable across different days and locations.  
For instance, the heavy traffic during the 7-8 AM rush hour consistently appears across days (\(A\) and \(B\)), and intersections with similar road structures often show comparable patterns(\(A\) and \(C\)).  
} 
}

Leveraging these patterns enables efficient dependency modeling across temporal and spatial dimensions.
By modeling the dependencies between segments and high-level events, representative patterns significantly reduce computational complexity while retaining long-range dependency information.
\blue{Additionally, combinations of representative patterns enable description of diverse high-level system events, endowing the model with the potential to generalize to unknown ones.}
This approach greatly lowers computational overhead. 
It preserves the ability to capture long-range dependencies while improving efficiency.  

More importantly, the number of representative patterns is determined by the high-level system events, ensuring that computational costs grow linearly with sequence length, effectively addressing computational bottlenecks.
Therefore, this approach demonstrates strong scalability for long-range dependency modeling in long and high-dimensional data.

\section{Overview of \sysname}
\label{sec:overview}
To leverage the representative segment patterns in time series data, we propose a two-phase method.
In the offline phase, representative segment patterns are extracted from the dataset. In the online phase, these patterns are used to accelerate long-range dependency modeling, enabling efficient forecasting with minimal computational overhead when handling large-scale time series data.

As shown in ~\figref{fig:overview}, the method operates in two phases: offline clustering and online adaptation.
\begin{itemize}
    \item In the \textit{offline} phase, the time series is first divided into smaller segments. 
    Next, these segments are assigned to buckets, where each bucket is associated with a specific prototype representing the common characteristics of the segments it contains. 
    Subsequently, these prototypes are iteratively refined to better capture the most representative local structures in the data.

    \item In the \textit{online} phase, the process starts by segmenting the incoming data and assigning each segment to the nearest prototype, resulting in the assignment matrix. 
    Next, another matrix is computed to capture the relationships between the input segments and the prototypes through a deep learning approach.
    By combining these two matrices, the method efficiently models long-range dependencies. 
    Importantly, this approach achieves linear computational complexity with respect to the input size, as the number of prototypes remains fixed regardless of the length of the online inputs.
    
    \item Furthermore, this approach is integrated into a dual-branch forecasting network, where long-range dependencies in the temporal and entity dimensions are modeled separately, features from both dimensions are extracted, and then fused to generate future predictions.
\end{itemize}

\section{Offline: Segment Clustering}
\label{sec:offline}

Understanding representative segment patterns in long and high-dimensional time series is crucial, as they provide a concise representation of complex dynamics. 
Those patterns and their corresponding high-level events often capture the system’s underlying structure, allowing for significant simplification during preprocessing. 
By leveraging these patterns, we can potentially reduce the need for exhaustive pairwise computations for segments in downstream global dependency modeling, alleviating computational overhead.

\blue{
We obtain representative segment patterns from the dataset by means of clustering.
Unlike the clustering approaches used at the channel-level~\cite{chen2025ccm, dong2024clustering, bandara2020clustering} in existing forecasting methods, this clustering method at the segment-level places more emphasis on the analysis of local similarity.
However, although the clustering method optimized with Euclidean distance is effective in finding simple patterns~\cite{zuo2023svp, zuo2024darker}, it struggles to identify the higher-order correlations that are crucial for dependency modeling.
}
This necessitates the use of more refined similarity metrics to account for the correlations between segments and guide the discovery of representative segment patterns.

\example{
\textit{
Suppose the traffic flow at intersection \( A \) changes as \(\{9, 10, 11\}\), at intersection \( B \) as \(\{7, 10, 13\}\), and at intersection \( C \) as \(\{11, 10, 9\}\). 
While the Euclidean distance between \( A \) and \( C \) is the same as that between \( A \) and \( B \), the flow changes at intersection \( A \) are significantly more correlated with those at intersection \( B \) than with intersection \( C \). 
This highlights the need to better differentiate between intersections \( B \) and \( C \) for modeling dependencies.
}
}

We address this problem by augmenting distance-based clustering objectives with correlation-based optimization. 
Let the input dataset \(\mathcal{D} = \{\mathcal{P}_e \mid e = 1, 2, \ldots, D\}\), where the sequence for each entity \(e\) is segmented into a sequence of segments \(\mathcal{P}_{e, i} \in \mathbb{R}^{p}\). Here, \(p\) represents the length of each segment, and each entity contributes \(T / p\) segments, assuming \(T\) is divisible by \(p\). 

A prototype set \(\mathcal{C} = \{\mathbf{c}_j \mid j = 1, 2, \ldots, k\}\) is defined, where each prototype \(\mathbf{c}_j \in \mathbb{R}^{p}\) encapsulates a representative segment pattern. 
The assignment of each segment \(\mathcal{P}_{e, i}\) to its closest prototype \(\mathbf{c}_{q_{e, i}}\) is determined by minimizing a composite loss that combines numerical similarity and correlation alignment:
\begin{equation}
\label{eq:assignment_eq}
q_{e, i} = \text{argmin}_{j} \left( \left\|\mathcal{P}_{e, i} - \mathbf{c}_j\right\|^2 + \alpha \cdot \big(1 - \text{corr}(\mathcal{P}_{e, i}, \mathbf{c}_j)\big) \right),
\end{equation}
where \(\text{corr}(\mathcal{P}_{e, i}, \mathbf{c}_j)\) denotes the Pearson correlation coefficient between \(\mathcal{P}_{e, i}\) and \(\mathbf{c}_j\), and \(\alpha\) is a fixed parameter balancing numerical similarity with correlation alignment. Each segment \(\mathcal{P}_{e, i}\) is then assigned to the corresponding bucket:
\begin{equation}
\mathcal{B}_j = \{\mathcal{P}_{e, i} \mid q_{e, i} = j\}.
\end{equation}

The prototype \(\mathbf{c}_j\) of each bucket is optimized based on the reconstruction loss and correlation loss. The reconstruction loss \(\mathcal{L}_{\text{rec}}\) enforces numerical consistency between each prototype and the mean characteristics of its assigned segments:
\begin{equation}
\mathcal{L}_{\text{rec}} = \sum_{j=1}^{k} \left\| \mathbf{c}_j - \text{mean}\left(\mathcal{B}_j\right) \right\|^2,
\end{equation}
where \(\text{mean}(\cdot)\) computes the mean segment across all \(\mathcal{P}_{e, i}\) assigned to prototype \(\mathbf{c}_j\). 
To capture temporal dynamics, the correlation loss \(\mathcal{L}_{\text{corr}}\) maximizes the alignment between prototypes and their assigned segments, using the Pearson correlation coefficient:
\begin{equation}
\label{eq:l_correlation}
\mathcal{L}_{\text{corr}} = -\sum_{j=1}^{k} \frac{1}{|\mathcal{B}_j|} \sum_{\mathcal{P}_{e, i} \in \mathcal{B}_j} \text{corr}(\mathcal{P}_{e, i}, \mathbf{c}_j).
\end{equation}

\blue{
For a dataset consisting of $L$ time steps, the computational complexity of both types of loss is on the order of $O(L)$.
}
The combined optimization objective is a weighted sum of the reconstruction and correlation losses:
\begin{equation}
\mathcal{L} = \mathcal{L}_{\text{rec}} + \alpha \cdot \mathcal{L}_{\text{corr}},
\end{equation}
where \(\alpha\) controls the weight of the correlation loss. 
The global objective integrates these levels of optimization:
\begin{equation}
\min_{\mathcal{C}} \mathcal{L}(\mathcal{C}, \mathbf{q}) 
\quad \text{subject to} \quad
\mathbf{q} \in \text{argmin}_{\mathbf{q}'} \mathcal{L}_{\text{assign}}(\mathbf{q}', \mathcal{C}),
\end{equation}
\noindent
\begin{equation}
\mathcal{L}_{\text{assign}}(\mathbf{q}', \mathcal{C}) = \sum_{e, i} \text{Dis}\big(\mathcal{P}_{e, i}, \mathbf{c}_{q'_{e, i}}\big),
\end{equation}
\noindent
and \(\text{Dis}(\mathcal{P}_{e, i}, \mathbf{c}_{q'_{e, i}})\) is defined as:
\begin{equation}
\text{Dis}(\mathcal{P}_{e, i}, \mathbf{c}_{q'_{e, i}}) = \left\| \mathcal{P}_{e, i} - \mathbf{c}_{q'_{e, i}} \right\|^2 + 
\alpha \cdot \big(1 - \text{corr}(\mathcal{P}_{e, i}, \mathbf{c}_{q'_{e, i}})\big).
\end{equation}

To solve this, we employ the AdamW optimizer~\cite{loshchilov2017adamw}, iteratively updating the prototype set \(\mathcal{C}\). This approach ensures that each prototype effectively balances numerical fidelity and morphological expressiveness.

The resulting prototypes serve as a compact yet expressive representation of the time series. 
This approach aligns well with applications requiring scalable analysis by embedding temporal relationships into the prototype learning process.

\begin{figure}[t]
    \centering
    \renewcommand{\algorithmicrequire}{\textbf{Input:}}
    \renewcommand{\algorithmicensure}{\textbf{Output:}}
    \begin{algorithm}[H]
        \caption{Segment Clustering}
        \label{alg:segmentation_prototype}
        \begin{algorithmic}[1]
            \REQUIRE Time series data \(\mathcal{D}\), Segment length \(p\), Number of prototypes \(k\)
            \ENSURE Updated set of prototypes \(\mathcal{C} = \{\mathbf{c}_1, \mathbf{c}_2, \ldots, \mathbf{c}_k\}\)\\
            \STATE Initialize the prototypes \(\mathcal{C}\) with \(k\) random prototypes
            \FOR{each entity \(e\) in \(\mathcal{D}\)}
                \STATE Segment the time series \(\mathcal{D}_e\) into segments of length \(p\)
                \STATE Store all segments in \(\mathcal{P}_e\)
            \ENDFOR
            \STATE Combine all segments: \(\mathcal{P} = \bigcup_{e=1}^{D} \mathcal{P}_e\)
            \WHILE{not converged}
                \FOR{each local pattern \(\mathcal{P}_{e, i} \in \mathcal{P}\)}
                    \STATE Assign \(\mathcal{P}_{e, i}\) to the nearest prototype \( \mathbf{c}_j \)
                    \STATE Place \(\mathcal{P}_{e, i}\) into the corresponding bucket \(b_j\)
                \ENDFOR
                \FOR{each bucket \(b_j \in \mathcal{B}\)}
                    \STATE Compute total loss \(\mathcal{L}_j\) for segments in \(b_j\)
                    \STATE Update prototype \( \mathbf{c}_j \) using gradient descent on \(\mathcal{L}_j\)
                \ENDFOR
            \ENDWHILE
            \RETURN the updated prototypes \(\mathcal{C}\)
        \end{algorithmic}
    \end{algorithm}
\end{figure}

\section{Online: Accelerated Dependency Modeling}
\label{sec:online}

\begin{figure}
  \centering
  \includegraphics[width=0.8\linewidth]{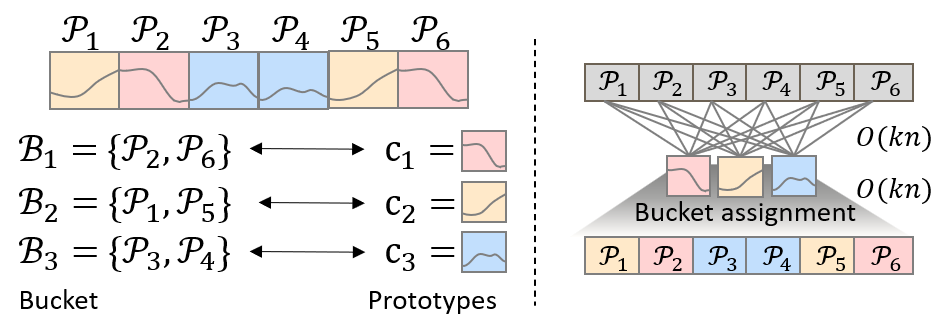}
  \caption{Online bucket assignment and the computation for dependency modeling.}
  \label{fig:online_calculations}
\end{figure}

In the online phase, our goal is to model long-range dependency between the temporal and entity dimensions of the input using observations over a specific time window containing \(N\) entities. 
To efficiently model long-range dependencies in multivariate time-series data, we propose Prototypes Attentive Modeling (ProtoAttn). 
As shown in \figref{fig:online_calculations}, ProtoAttn leverages prototypes derived in the offline phase to group input sequences into segment buckets and models interactions between these prototypes and the original segments. This approach avoids the quadratic complexity of traditional self-attention mechanisms while preserving critical temporal relationships.

\subsection{Implementation of ProtoAttn}
The input sequence undergoes the same segmentation process as in the offline phase, resulting in \blue{\(l \times N\)} input segments, where \blue{\(l = L / p\)} represents the number of temporal segments per entity, and \(N\) is the total number of entities. Without loss of generality, we assume that \(p\) divides \(L\) evenly. 
For simplicity, we focus on modeling \(n\) segments which is denoted as \(\mathcal{P}\), irrespective of whether they originate from different timestamps of the same entity or from different entities at the same timestamp.
Let \(k\) denote the number of prototypes derived from the offline phase. ProtoAttn starts by computing a mapping matrix \(A \in \mathbb{R}^{l \times k}\) that assigns input segments to the pattern buckets established offline, linking the input data to the prototypes. This mapping allows the transformation of input sequences into Query, Key, and Value:
\begin{equation}
    \mathcal{C}_Q = \mathcal{C} W_E, \quad K =  \mathcal{P}W_K, \quad V = \mathcal{P}W_V
\end{equation}
where \(\mathcal{P} \in \mathbb{R}^{l \times p}\) represents the input time segments, \(\mathcal{C} \in \mathbb{R}^{k \times p}\) represents the prototypes, and \(W_E, W_K, W_V \in \mathbb{R}^{p \times d}\) are the learnable projection matrices for feature mapping.

Next, the query matrix \(Q\) is constructed, where each query vector \(q_i \in \mathbb{R}^d\) is represented by its corresponding cluster assignment. 
This allows us to define the approximate representation of \(q_i\) as:
\begin{equation}
    \hat{q}_i = A_i \mathcal{C}_Q
\end{equation}
where \( \mathcal{C}_Q \in \mathbb{R}^{k \times d} \) represents a matrix of embedded prototypes, and \( A_i \in \mathbb{R}^{1 \times k} \) is a one-hot vector denoting the cluster assignment for \( q_i \).
To compute the attention weights for each centroid \( c_{q_i} \) with respect to the keys \( K \), we perform a dot product followed by a softmax normalization:
\begin{equation}
    \alpha_i = \text{softmax} \left( \frac{c_{q_i}^\top K^\top}{\sqrt{d_k}} \right)
\end{equation}
where \( \alpha_i \in \mathbb{R}^n \) represents the attention distribution for the centroid \( c_{q_i} \).
The final output of the ProtoAttn mechanism is computed as a weighted sum of the value matrix \( V \) using the attention weights \( \alpha_i \):
\begin{equation}
    \text{ProtoAttn}(q, K, V) = \alpha_i V
\end{equation}
expanding this across all query centroids yields will get us:
\begin{equation}
    \text{ProtoAttn}(\mathcal{C}_Q, K, V) = A \left( \text{softmax} \left( \frac{\mathcal{C}_Q K^\top}{\sqrt{d_k}} \right) \right) V
\end{equation}

This formulation aggregates attention outputs for each centroid in the query space, ensuring that queries sharing the same centroid \( c_{q_i} \) will yield identical attention weights:
\begin{equation}
    A_i = A_j \implies \alpha_i = \alpha_j
\end{equation}



\begin{figure}[t]
    \centering
    \renewcommand{\algorithmicrequire}{\textbf{Input:}}
    \renewcommand{\algorithmicensure}{\textbf{Output:}}
    \begin{algorithm}[H]
        \caption{Online Dependency Modeling}
        \label{alg:ProtoAttn_matrix}
        \begin{algorithmic}[1]
            \REQUIRE Input segment matrix \( \mathcal{P} \in \mathbb{R}^{l \times d} \), Prototypes Set \( C \)
            \ENSURE Attention output \( \text{ProtoAttn}(Q, K, V) \)

            \STATE Initialize \( A \in \mathbb{R}^{l \times k} \) as a zero matrix.
            \FOR{each segment \( \mathcal{P}_i \) in \( \mathcal{P} \)}
                \STATE Assign \( A[i, j] \leftarrow 1 \), where \( j \) is determined as the closest prototype to \( \mathcal{P}_i \) by \eqref{eq:assignment_eq}.
            \ENDFOR

            \STATE \( K \leftarrow \mathcal{P}W_K \in \mathbb{R}^{l \times d} \)
            \STATE \( V \leftarrow \mathcal{P}W_Q \in \mathbb{R}^{l \times d} \)

            \STATE \( \alpha \leftarrow \text{softmax}\left(\frac{\mathcal{C}_Q K^\top}{\sqrt{d_k}}\right) \) 

            \STATE \( \text{ProtoAttn}(\mathcal{C}_Q, K, V) \leftarrow \alpha V \) 

            \STATE \( \text{Output} \leftarrow A \times \text{ProtoAttn}(Q, K, V) \) 

            \RETURN \( \text{Output} \)
        \end{algorithmic}
    \end{algorithm}
\end{figure}

\subsection{Analysis of the Online Dependency Modeling}
\label{sub:online_complexity_analysis}
The pseudocode in \algref{alg:ProtoAttn_matrix} demonstrates the matrix computation form of the above process, which can better leverage the capabilities of parallel processors.

\fakeparagraph{Complexity Analysis}
\blue{
The complexity analysis for the ProtoAttn begins with the computation of the assignment matrix \( A \), where each segment \( \mathcal{P}_i \) is assigned to the nearest prototype among \( k \) candidates, with a complexity of \( \mathcal{O}(l \cdot k \cdot d) \). Constructing the Query, Key, and Value matrices involves projecting the prototypes \( \mathcal{C} \) and segments \( \mathcal{P} \), resulting in a total complexity of \( \mathcal{O}(k \cdot d^2 + l \cdot d^2) \). Computing attention weights \( \alpha \), applying them to the Value matrix \( V \), and mapping outputs back to segments together have a complexity of \( \mathcal{O}(k \cdot l \cdot d + l \cdot k \cdot d) \). Summing these, the total complexity is \( \mathcal{O}(l \cdot (k \cdot d + d^2) + k \cdot d^2) \).
Since \( d \), and \( p \) are constants, this simplifies to \( \mathcal{O}(kl) \), making ProtoAttn highly efficient with linear complexity relative to the number of segments \(l\).
}

\fakeparagraph{Approximation Analysis of ProtoAttn}
An important property in the above process is that the number of fixed patterns discovered from the entire dataset is identified during the offline stage, and it is independent of the length of historical data input during the online stage or the length of the predicted future. 
Formally, when the input data is divided into segments, its \textit{rank} should be smaller than the number of fixed patterns discovered from the entire dataset, \ie the low-rank nature of the input data.
We provide a theoretical proof that, when the input data indeed exhibits low-rank properties, the aforementioned process can be approximated as the long-range dependencies in self-attention process.

\theorem{
Let \( \mathcal{P} \in \mathbb{R}^{l \times p} \) be an input sequence matrix composed of segments, with \(\text{rank}(\mathcal{P}) \leq r\), \(r\) is the number of representative segment patterns found offline in the whole dataset, and let \( W^Q, W^K \in \mathbb{R}^{d \times d} \) be weight matrices. Let \( w \) be a vector drawn from any column of the product \( (W_Q W_K^T\). For any \( \epsilon \in (0,1) \), there exists a low-rank matrix approximation \( \tilde{\mathcal{P}} = AC \), where \( A \in \mathbb{R}^{l \times k} \) and \( C \in \mathbb{R}^{k \times d} \), \(k\) is the number of patterns in online input, which satisfies \(k \leq r\), such that the following inequality holds:
\begin{equation}
\label{eq:theorem1}
    \Pr\left(\|\tilde{\mathcal{P}} w^T - \mathcal{P} w^T\| \leq \epsilon \|\mathcal{P} w^T\|\right) > 1 - o(1),
\end{equation}
where the dimension \( k \) satisfies:
\begin{equation}
k = O\left(\frac{\log r}{\epsilon^2}\right).
\end{equation}
}

\proof{
Since \( \mathcal{P} \) has rank at most \( r \), we can decompose it as \( \mathcal{P} = UV \), where \( U \in \mathbb{R}^{l \times r} \) is a matrix with orthonormal columns and \( V \in \mathbb{R}^{r \times p} \).
We aim to show that for a matrix \( \tilde{\mathcal{P}} = AC \), the low-rank approximation error with respect to \( w^T \) is bounded with high probability. 

The first step is to show that we can approximate the matrix \( V \) using a matrix of rank \( k \). To achieve this, we define \( A' \in \mathbb{R}^{r \times k} \) and \( C \in \mathbb{R}^{k \times d} \), and let \(\tilde{V} = A'C.\).
This ensures that \( \text{rank}(\tilde{V}) \leq \text{rank}(A') = k \leq r \), so \( \tilde{V} \) is a low-rank approximation of \( V \). 
Our goal is to show that \( \tilde{V} \) approximates \( V \) well, i.e., we want to prove that the approximation error of \( V \) with respect to \( \tilde{V} \) is small with high probability.
\begin{equation}
    \Pr\left(\|\tilde{V} w^T - V w^T\| \leq \epsilon \|V w^T\|\right) > 1 - o(1),
\end{equation}

Here we provide a constructive proof by making \( A' = VC^T \). 
We aim to show:
\begin{equation}
\label{eq:v_appr}
    \Pr\left(\| VC^T C w^T - V w^T \| \leq \epsilon \|V w^T\|\right) > 1 - o(1).
\end{equation}

Using the Johnson-Lindenstrauss (JL) lemma~\cite{wang2020linformer}, for any matrix \( R \in \mathbb{R}^{k \times d} \) and vector \( x \in \mathbb{R}^k \), we have the following inequality for any \( y \in \mathbb{R}^k \):
\begin{equation}
\label{eq:jl-bound}
    \Pr\left(\| x R^T R y^T - xy^T \| \leq \epsilon \|xy^T\|\right) > 1 - 2e^{-(\epsilon^2 - \epsilon^3)k / 4},
\end{equation}
where \( k \) satisfies:
\begin{equation}
k = \frac{5 \log r}{\epsilon^2 - \epsilon^3}.
\end{equation}

To apply this bound to the original matrix \( V \), we define:
\begin{equation}
    M_1 = VC^T Cw^T - Vw^T.
\end{equation}
\begin{equation}
    M_2 = x C^T C w^T - xw^T,
\end{equation}
\begin{equation}
\begin{split}
        \Pr \left(\| M_1 \| \leq \epsilon \| V w^T \| \right)
        &\geq 1 - \sum_{x \in V} \Pr \left( \| M_2 \| \leq \epsilon \|x w^T \| \right) \\
        &\geq 1 - 2n e^{-(\epsilon^2 - \epsilon^3) k / 4}
\end{split}
\end{equation}

Now we finished proving \eqref{eq:v_appr}.
Next, we can observe that the error between \( \tilde{\mathcal{P}}w^T \) and \( \mathcal{P}w^T \) can be written as:
\begin{equation}
    \|\tilde{\mathcal{P}}w^T - \mathcal{P}w^T\| = \|U(A'Cw^T - Vw^T)\|.
\end{equation}

Using the fact that \( U \) is a matrix with orthonormal columns, we know that:
\begin{equation}
    \|U(A'Cw^T - Vw^T)\| = \|A'Cw^T - Vw^T\|.
\end{equation}

Since \( \mathcal{P}w^T = UVw^T \), we can conclude that:
\begin{equation}
    \Pr\left( \|\tilde{\mathcal{P}}w^T - \mathcal{P}w^T \| \leq \epsilon \|\mathcal{P}w^T\| \right) > 1 - o(1),
\end{equation}

Thus, let \(A = UA'\), we have shown that the low-rank approximation \( ACw^T \) of \( \mathcal{P}w^T \) has a small error with high probability, completing the proof.

\hfill $\square$
}

This theorem demonstrates that the input sequence matrix can be decomposed into a low-rank form that effectively approximates the original attention matrix without introducing bias. Notably, the rank of the approximation is independent of the input size and is determined by the intrinsic properties of the time series itself, as discussed in \secref{sec:motivation}.



\section{Dual-Branch Forecasting}
\label{sec:dual-branch}

In this section, we present the design of the dual-branch fusion forecasting network, \sysname, which utilizes the former two-phase process to model temporal and entity correlations for accurate foreacasting. 
In \secref{sub:dependency_modeling}, we shows how we use the online operation to construct the dual-branch feature extractor.
In \secref{sub:fusion}, we introduce a readout mechanism to capture useful information from extracted features efficiently to enhance forecasting accuracy.

\subsection{Efficient Long-range Dependency Extractor}
\label{sub:dependency_modeling}

Accurate time series forecasting requires capturing both temporal dynamics and inter-entity relationships. 
The temporal dimension represents sequential dependencies that evolve over time, while the entity dimension models the interactions between different variables, which is particularly crucial in multivariate forecasting scenarios.

Unlike traditional approaches that rely on multiple stacked layers to model dependencies, we propose a dual-branch design that simultaneously captures temporal and entity correlations within a unified framework.  
This design integrates seamlessly with our accelerated correlation modeling in the online phase, significantly enhancing both the expressiveness and efficiency of forecasting models.

To capture temporal dependencies, we construct long-range relationships along the time dimension by processing input data sequentially.  
This is achieved by modeling the interactions and dependencies between segments at different positions within the same entity, which effectively encapsulates the sequence's intrinsic temporal characteristics. 
To capture the significant impact of entity interactions on system behavior, we shift the focus to the entity dimension, enabling the model to learn entity-level relationships at each time point.
The procedure for feature extraction in both dimensions is described in \algref{alg:dual_branch_extraction}, where the temporal and entity feature matrices, \( \mathcal{H}_t \) and \( \mathcal{H}_e \), are computed using parallel operations on the input time series data.

\begin{figure}[t]
    \centering
    \renewcommand{\algorithmicrequire}{\textbf{Input:}}
    \renewcommand{\algorithmicensure}{\textbf{Output:}}
    \begin{algorithm}[H]
        \caption{Dual-Branch Feature Extraction Network}
        \label{alg:dual_branch_extraction}
        \begin{algorithmic}[1]
            \REQUIRE Input time series data \(\mathcal{X}\) with dimensions \([L \times N]\), segment length \(p\)
            \ENSURE Temporal features \(\mathcal{H}_{t}\) and entity features \(\mathcal{H}_{e}\)
            
            \STATE Divide \(\mathcal{X}\) along the time dimension into overlapping/non-overlapping segments of length \(p\), forming \(\mathcal{P}\)

            \FOR{\textbf{each entity} \(i = 1\) to \(N\)}
                \STATE Select the time segments \(\mathcal{P}_{t}^{(i)}\) for entity \(i\)
                \STATE Compute temporal features:
                \[
                \mathcal{H}_{t}^{(i)} \leftarrow \text{LayerNorm}\left(\text{OnlineModeling}(\mathcal{P}_{t}^{(i)}) + \mathcal{P}_{t}^{(i)}\right)
                \]
            \ENDFOR
            \STATE Concatenate \(\mathcal{H}_{t}^{(i)}\) across all entities along the entity dimension to form \(\mathcal{H}_{t}\)

            \FOR{\textbf{each time segment} \(j = 1\) to \(L / p\)}
                \STATE Select the entity segments \(\mathcal{P}_{e}^{(j)}\) for time segment \(j\)
                \STATE Compute entity features:
                \[
                \mathcal{H}_{e}^{(j)} \leftarrow \text{LayerNorm}\left(\text{OnlineModeling}(\mathcal{P}_{e}^{(j)}) + \mathcal{P}_{e}^{(j)}\right)
                \]
            \ENDFOR
            \STATE Concatenate \(\mathcal{H}_{e}^{(j)}\) across all time segments along the time dimension to form \(\mathcal{H}_{e}\)
            
            \RETURN \(\mathcal{H}_{t}, \mathcal{H}_{e}\)
        \end{algorithmic}
    \end{algorithm}
\end{figure}

            



\subsection{Efficient Parallel Fusion Module}
\label{sub:fusion}

\begin{figure}
    \centering
    \includegraphics[width=0.8\linewidth]{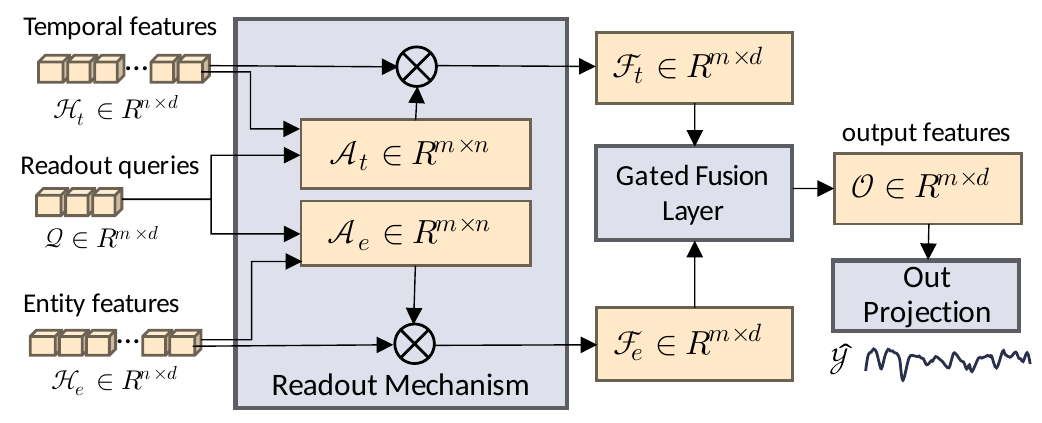}
    \caption{The information flow of Parallel Fusion Module.}
    \label{fig:fusion_layer}
\end{figure}

Efficiently integrating diverse feature sets is crucial for enhancing predictive performance in sequential data tasks. 
As shown in \figref{fig:fusion_layer}, our proposed \textit{Parallel Fusion Module} merges temporal and entity features in a computationally efficient manner, achieving linear scalability with respect to input sequence length. 

The process begins by generating a fixed number of \textit{readout queries} from the input features, with their length \(m\) corresponding to the desired forecast horizon.  
Using these queries, we compute their correlations with extracted features to capture their dependencies, producing prediction representations enhanced by each dimension.  
The gating mechanism dynamically adjusts the contributions of temporal and entity features, prioritizing the most relevant information for accurate forecasting.
Finally, the output is directly projected to produce the future prediction.

The pseudo-code of the Parallel Fusion Module is presented in \algref{alg:feature_fusion}.
The size of correlation matrices are linear to the input length as the number of readout queries are fixed.

\begin{figure}[t]
    \centering
    \renewcommand{\algorithmicrequire}{\textbf{Input:}}
    \renewcommand{\algorithmicensure}{\textbf{Output:}}
    \begin{algorithm}[H]
        \caption{Parallel Fusion Module}
        \label{alg:feature_fusion}
        \begin{algorithmic}[1]
            \REQUIRE Temporal features \(\mathcal{H}_{t}\), Entity features \(\mathcal{H}_{e}\), Number of readout queries \(m\)
            \ENSURE Forecasting result \(\hat{\mathcal{Y}}\)

            \STATE Project input features into \(m\) fixed readout queries \(\mathcal{Q}\) to represent essential information

            \STATE Compute \(\mathcal{A}_{t} = \text{softmax}\left(\frac{\mathcal{H}_{t} \mathcal{Q}^\top}{\sqrt{d}}\right)\) for temporal features
            \STATE Compute \(\mathcal{A}_{e} = \text{softmax}\left(\frac{\mathcal{H}_{e} \mathcal{Q}^\top}{\sqrt{d}}\right)\) for entity features

            \STATE Calculate \(\mathcal{F}_{t} = \mathcal{A}_{t} \mathcal{H}_{t}\) and \(\mathcal{F}_{e} = \mathcal{A}_{e} \mathcal{H}_{e}\)

            \STATE Concatenate \(\mathcal{F}_{t}\) and \(\mathcal{F}_{e}\) to obtain \(\mathcal{F}_{\text{proj}}\)
            \STATE Use a gating mechanism \(g(\mathcal{F}_{\text{proj}})\) to adjust the contributions of each features:
            \STATE \(\mathcal{O} \leftarrow g(\mathcal{F}_{\text{proj}}) \odot \mathcal{F}_{t} + (1 - g(\mathcal{F}_{\text{proj}})) \odot \mathcal{F}_{e}\)

            \STATE Map the final output \(\mathcal{O}\) to the forecasted result \(\hat{\mathcal{Y}}\)
            \RETURN \(\hat{\mathcal{Y}}\)
        \end{algorithmic}
    \end{algorithm}
\end{figure}

\fakeparagraph{Complexity Analysis}
\blue{
The overall complexity of \sysname is mainly determined by online dependency modeling.
For temporal feature extraction, processing the sequence along the temporal dimension divides it into \(l = L/p\) segments, with a complexity of \(O(kL/p)\) (\(k\) is the number of prototypes, \(L\) is the temporal length). For entity feature extraction, modeling dependencies between \(N\) entities at each time step gives \(N\) segments and a complexity of \(O(kN)\).

In the fusion processing, projecting input features \(\mathcal{H}_t\in\mathbb{R}^{l\times d}\) and \(\mathcal{H}_e\in\mathbb{R}^{N\times d}\) into \(m\) fixed readout queries has a complexity of \(\mathcal{O}(m\cdot l\cdot d + m\cdot N\cdot d)\). Computing correlation matrices \(\mathcal{A}_t\) and \(\mathcal{A}_e\) also costs \(\mathcal{O}(m\cdot l\cdot d + m\cdot N\cdot d)\), and aggregating features via matrix multiplications is \(\mathcal{O}(m\cdot l\cdot d + m\cdot N\cdot d)\), while the fusion projection and gating mechanism add \(\mathcal{O}(m\cdot d^2)\). Summing these, the total complexity is \(\mathcal{O}(m\cdot l\cdot d + m\cdot N\cdot d + m\cdot d^2)\). With \(m\) fixed, it simplifies to \(\mathcal{O}(l\cdot d + N\cdot d + d^2)\), and for large-scale inputs, further to \(\mathcal{O}(l\cdot d + N\cdot d)\). Since \(l=(L/p)\), the complexity is \(\mathcal{O}((L/p)\cdot d + N\cdot d)\), showing linear scalability with \(L\) and \(N\).

}

\section{Experiments}
\label{sec:experiments}
In this section, we evaluate \sysname on the long-term multivariate forecasting task, with a primary focus on accuracy and efficiency comparisons. 
We include many widely-used datasets from various real-world scenarios, including traffic, weather, and power systems. 
\blue{Our code is now publicly available\footnotemark.}

\footnotetext{\blue{https://anonymous.4open.science/r/ICDE25-ScalableForecasting-0AC5/}}

\subsection{Long-range Forecasting Experiment}
\label{sub:forecasting}

\fakeparagraph{Datasets}
We evaluate our proposed \sysname on multiple real-world multivariate time series (MTS) datasets, as summarized in \tabref{tab:dataset_details}. Following standard data split configurations, we use a 7/1/2 train/validation/test ratio for Weather, Electricity, and Traffic datasets, and a 6/2/2 ratio for ETT, PEMS04, and PEMS08 datasets. These datasets are normalized using statistical information derived from the training set, consistent with prior studies~\cite{nie2023patchtst, lai2023lightcts}. 
Widely recognized as benchmarks in MTS forecasting research~\cite{wu2021autoformer, wu2023timesnet, zeng2023dlinear, nie2023patchtst, zhang2023crossformer}, these datasets span a variety of sizes and entity counts, representing diverse real-world applications and effectively capturing the multifaceted characteristics of multivariate time series data.

\begin{table}[h!]
\centering
\scriptsize
\caption{Statistics of multivariate time series datasets.}
\label{tab:dataset_details}
\setlength{\tabcolsep}{5pt} 
\setlength{\extrarowheight}{-1pt} 
\begin{tabular}{l l l l l l p{1cm}}
\hline
\textbf{Dataset} & \textbf{Domain} & \textbf{Frequency} & \textbf{Lengths} & \textbf{Dim} & \textbf{Split}\\ 

\hline \hline
PEMS04 & Traffic & 5 mins & 16,992 & 307 & 6:2:2 \\ 
PEMS08 & Traffic & 5 mins & 17,856 & 170 & 6:2:2 \\ 
ETTh1 & Electricity & 1 hour & 14,400 & 7 & 6:2:2 \\ 
ETTm1 & Electricity & 15 mins & 57,600 & 7 & 6:2:2 \\ 
Traffic & Traffic & 1 hour & 17,544 & 862 & 7:1:2 \\ 
Electricity & Electricity & 1 hour & 26,304 & 321 & 7:1:2  \\
Weather & Environment & 10 mins & 52,696 & 21 & 7:1:2 \\ 
\hline
\end{tabular}
\end{table}

\begin{table*}[ht]
  \centering
  \caption{Comparison of long-range forecasting accuracy with baselines}
  \label{tab:main_forecasting_results}
  \footnotesize
  \setlength{\tabcolsep}{2.2pt} 
  \setlength{\extrarowheight}{-0.5pt} 
  \renewcommand{\arraystretch}{1.1}
  \begin{tabular}{cc|cc|cc|cc|cc|cc|cc|cc|cc}
    \hline
    \multicolumn{2}{c|}{\textbf{Models}} & \multicolumn{2}{c|}{\sysname} & \multicolumn{2}{c|}{PatchTST} & \multicolumn{2}{c|}{Crossformer} & \multicolumn{2}{c|}{MTGNN} & \multicolumn{2}{c|}{Graph Wavenet} & \multicolumn{2}{c|}{TimesNet} & \multicolumn{2}{c|}{LightCTS} & \multicolumn{2}{c}{DLinear}\\
    \hline
    \multicolumn{2}{c|}{\textbf{Metrics}} & MSE & MAE & MSE & MAE & MSE & MAE & MSE & MAE & MSE & MAE & MSE & MAE & MSE & MAE & MSE & MAE \\
    \hline
    \hline
    \multirow{2}{*}{PEMS04}
    & 96 & \textbf{0.0758} & \textbf{0.170} & 0.102 & 0.228 & 0.103 & 0.209 & \underline{0.0835} & 0.189 & 0.0838 & \underline{0.186} & 0.0950 & 0.203 & 0.115 & 0.229 & 0.129 & 0.219\\
    & 336 & \textbf{0.0936} & \textbf{0.190} & 0.120 & 0.240 & 0.125 & 0.234 & 0.108 & 0.218 & 0.105 & 0.209 & \underline{0.101} & \underline{0.208} & 0.123 & 0.230 & 0.161 & 0.245\\
    \hline
    \multirow{2}{*}{PEMS08}
    & 96 & \textbf{0.0504} & \textbf{0.139} & 0.0775 & 0.200 & 0.0773 & 0.190 & \underline{0.0581} & \underline{0.158} & 0.0615 & 0.157 & 0.66 & 0.172 & 0.0835 & 0.196 & 0.115 & 0.207\\
    & 336 & \textbf{0.066} & \textbf{0.161} & 0.0833 & 0.201 & 0.0929 & 0.207 & 0.0757 & 0.184 & 0.0793 & 0.183 & \underline{0.07} & \underline{0.177} & 0.0900 & 0.200 & 0.140 & 0.233\\
    \hline
    \multirow{2}{*}{ETTh1}
    & 96 & \textbf{0.372} & \textbf{0.402} & 0.391 & 0.422 & 0.418 & 0.449 & 0.454 & 0.472 & 0.458 & 0.472 & 0.428 & 0.452 & 0.401 & 0.429 & 0.401 & 0.424 \\
    & 336 &\textbf{0.391} & \textbf{0.423} & 0.434 & 0.463 & 0.751 & 0.681 & 0.457 & 0.474 & 0.543 & 0.531 & 0.454 & \underline{0.425} & 0.506 & 0.495 & \underline{0.424} & 0.446\\
    \hline
    \multirow{2}{*}{ETTm1}
    & 96 & \underline{0.304} & \textbf{0.352} & \textbf{0.297} & \underline{0.354} & 0.324 & 0.370 & 0.366 & 0.427 & 0.358 & 0.402 & 0.316 & 0.369 & 0.312 & 0.363 & 0.307 & 0.358\\
    & 336 & \textbf{0.366} & \textbf{0.394} & \underline{0.368} & \underline{0.395} & 0.418 & 0.442 & 0.523 & 0.525 & 0.449 & 0.459 & 0.381 & 0.410 & 0.392 & 0.417 & 0.371 & 0.399\\
    \hline
    \multirow{2}{*}{Traffic}
    & 96 & \textbf{0.387} & \textbf{0.275} & \underline{0.388} & 0.299 & 0.520 & \underline{0.287} & 0.492 & 0.291 & 0.523 & 0.292 & 0.623 & 0.337 & 0.596 & 0.415 & 0.395 & 0.276\\
    & 336 & \textbf{0.414} & \textbf{0.285} & \underline{0.418} & 0.313 & 0.525 & \underline{0.286} & 0.535 & 0.321 & 0.570 & 0.325 & 0.652 & 0.374 & 0.623 & 0.383 & 0.421 & 0.331\\
    \hline
    \multirow{2}{*}{Electricity}
    & 96 & \textbf{0.132} & \textbf{0.227} & 0.155 & 0.274 & 0.136 & 0.236 & 0.137 & 0.238 & 0.146 & 0.247 & 0.190 & 0.296 & 0.177 & 0.279 & \underline{0.135} & \underline{0.233}\\
    & 336 & \textbf{0.164} & \textbf{0.259} & 0.192 & 0.305 & \underline{0.166} & \underline{0.266} & 0.176 & 0.281 & 0.191 & 0.292 & 0.206 & 0.309 & 0.208 & 0.307 & 0.166 & 0.267\\
    \hline
    \multirow{2}{*}{Weather}
    & 96 & \textbf{0.148} & \textbf{0.199 }& 0.152 & 0.205 & 0.154 & 0.220 & 0.1613 & 0.227 & 0.161 & 0.221 & 0.170 & 0.229 & \underline{0.149} & \underline{0.203} & 0.155 & 0.223\\
    & 336 & \underline{0.236} & \textbf{0.280} & 0.240 & 0.284 & 0.264 & 0.333 & 0.260 & 0.330 & 0.259 & 0.323 & 0.260 & 0.305 & \textbf{0.234} & \underline{0.281} & 0.239 & 0.302\\
    \hline
  \end{tabular}
\end{table*}

\begin{figure*}[ht]
    \centering
    \includegraphics[width=0.8\linewidth]{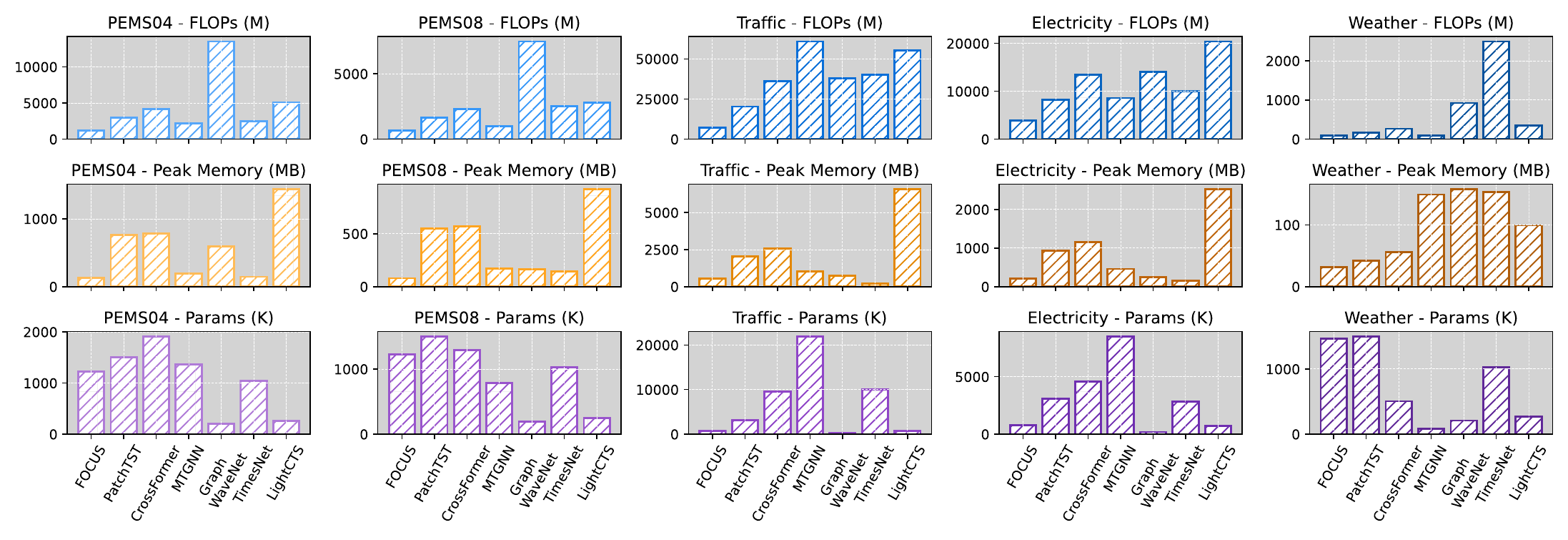}
    \caption{Comparison of FLOPs, Peak Memory Occupation, Number of Parameters with baselines.}
    \label{fig:efficiency_comparison}
\end{figure*}

\fakeparagraph{Metrics}
In terms of evaluation metrics, we focus primarily on the predictive accuracy and computational efficiency of the model.  
For accuracy, we utilize Mean Absolute Error (MAE) and Mean Squared Error (MSE), where lower values indicate better forecasting performance. These metrics are widely recognized as benchmarks for evaluating the accuracy of time series forecasting, as both quantify the numerical differences between the predicted and actual sequences.  
For efficiency, in alignment with existing conventions~\cite{lai2023lightcts} and to minimize the impact of varying deep learning platforms and operating system conditions (\eg concurrent program execution), we adopt three key metrics: the number of floating-point operations (FLOPs) involved during inference, peak memory usage during inference (Peak Memory), and the model’s parameter count. 
FLOPs reflect the total computational workload of the model, peak memory usage represents the space required for intermediate computation results, and parameter count indicates the storage capacity demands. 
These metrics collectively provide a comprehensive assessment of the computational bottlenecks that time series forecasting models might face in real-world scenarios.

\fakeparagraph{Baselines}
We compare \sysname with state-of-the-art models for multivariate time series forecasting, encompassing various architectures as follows:
\begin{itemize} 
\item PatchTST\cite{nie2023patchtst}: A single-channel forecasting model based on a transformer architecture, widely used for long-term forecasting benchmarks and known for its robust and consistent accuracy across diverse datasets. 
\item Crossformer\cite{zhang2023crossformer}: A sophisticated model that employs Transformer structures along both the temporal and entity dimensions for enhanced performance. 
\item MTGNN\cite{wu2020mtgnn}: A widely adopted model that integrates adaptive graph convolutional networks with temporal convolutional networks in an optimized architecture. 
\item Graph Wavenet\cite{wu2019graphwavenet}: An efficient forecasting model using adaptive graph modeling and dilated causal convolution. 
\item TimesNet\cite{wu2023timesnet}: A model that introduces Temporal 2D-Variation Modeling to achieve strong performance. 
\item LightCTS\cite{lai2023lightcts}: A recent model that employs a refined structural design for efficient time series forecasting. 
\item DLinear\cite{zeng2023dlinear}: A simple yet effective model based on linear layers for multi-step prediction.
\end{itemize}

\fakeparagraph{Implementation Details}
All experiments are conducted on a cloud server equipped with two NVIDIA Tesla V100 GPUs, each with 32GB of memory. To ensure fairness, we use the original configurations for all baseline models. To align with existing works in long-term forecasting ~\cite{nie2023patchtst, zhang2023crossformer}, a lookback window of 512 steps is adopted, with forecasting horizons set to 96 and 336 steps. 
The correlation loss weight is set to \(\alpha = 0.2\) during the offline phase. 
For \sysname, we use a single-layer structure for both the Temporal Extractor and the Entity Extractor, while the feature mixing layer employs readout tokens. 
The number of readout tokens (\(m\)) is set to 6 for a forecasting horizon of 96 steps and 21 for 336 steps.
The embedding size (\(d\)) is configured as \(d = 128\) for the PEMS04 and PEMS08 datasets, and \(d = 64\) for all other datasets.
\blue{Other hyperparameters employed in the experiment, including the segment length $p$ and the number of prototypes $k$, were obtained through the grid-search method.}

\fakeparagraph{Conclusion}
\label{sub:forecasting_results}
The results in \tabref{tab:main_forecasting_results} highlight \sysname's outstanding accuracy in long-term forecasting across seven diverse datasets.
The best results are in \textbf{bold}, and the second-best results are \underline{underlined}.
\sysname consistently outperforms nearly all baseline models in both MAE and MSE metrics, with only a marginal MSE difference in favor of LightCTS on the Weather dataset and a marginal difference with PatchTST on ETTm1. 
In most datasets and scenarios, \sysname demonstrates a significant advantage, surpassing models like PatchTST and Crossformer with superior accuracy. 
On more challenging datasets, such as PEMS04, PEMS08, and Traffic, \sysname exhibits both high precision and an efficient approach.

As illustrated in ~\figref{fig:efficiency_comparison}, \sysname demonstrates significantly lower FLOPs and peak memory consumption compared to other baseline models when processing long input sequences. This efficiency ensures that \sysname remains practical even on resource-constrained devices, enabling robust performance without overburdening hardware. While \sysname may exhibit slightly higher parameter counts in specific scenarios, it effectively mitigates peak memory usage by minimizing intermediate variable storage, maintaining a balance between computational cost and accuracy.

\subsection{Parameter Study}
\label{sub:parameter_study}
\begin{figure}[t]
  \centering
  \subfloat[]
  {
      \label{fig:param_study_k}\includegraphics[width=0.22\linewidth]{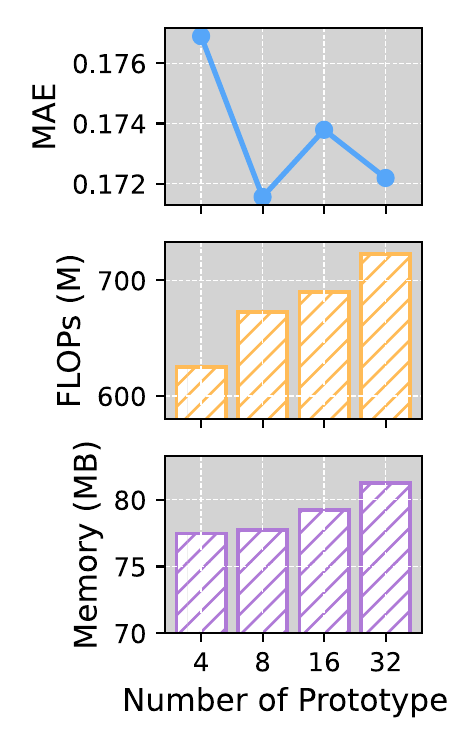}
    }
    \hfill
  \subfloat[]
  {
      \label{fig:param_study_d}\includegraphics[width=0.22\linewidth]{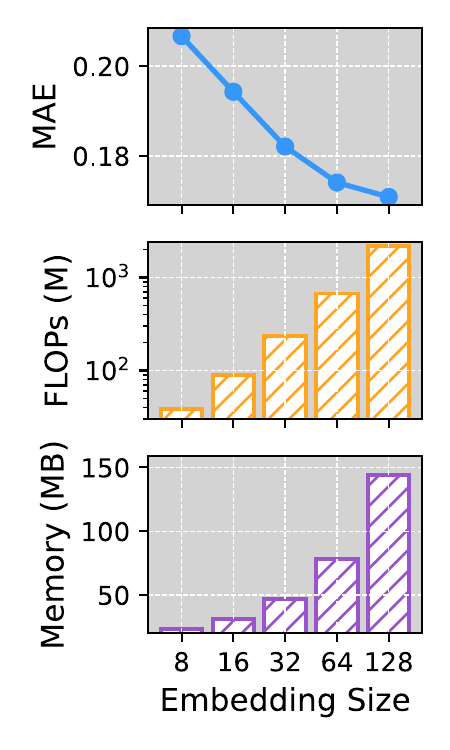}
  }
  \hfill
  \subfloat[]
  {
      \label{fig:param_study_l}\includegraphics[width=0.22\linewidth]{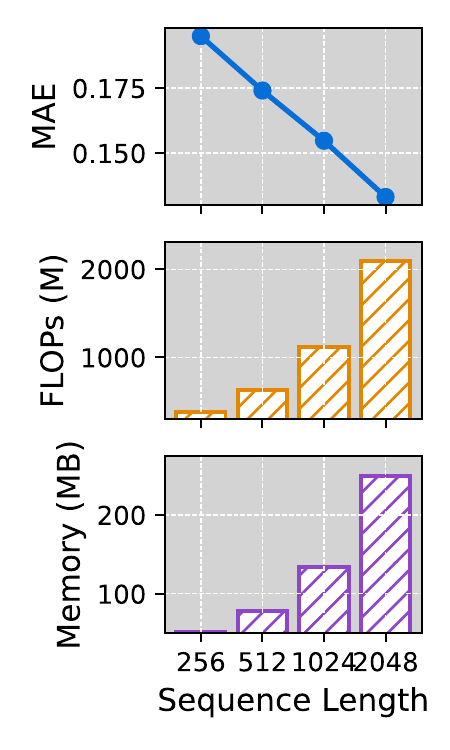}
    }   
\subfloat[]
  {
      \label{fig:param_study_p}\includegraphics[width=0.22\linewidth]{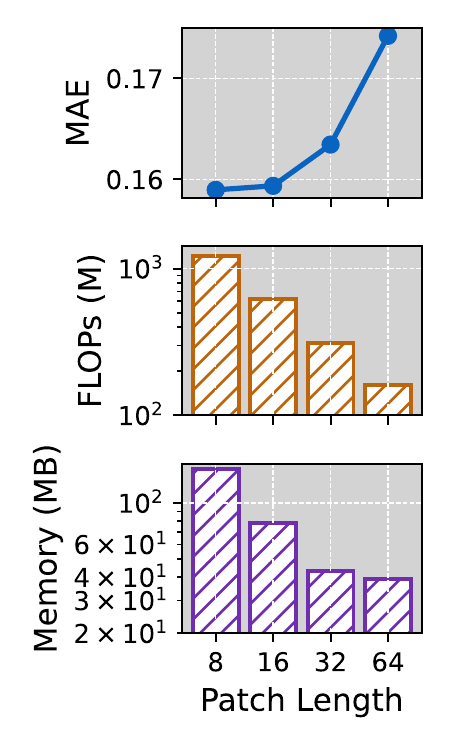}
    }
  \caption{Impact of (a) number of prototypes \(k\), (b) embedding size \(d\), (c) length of the input sequence \(L\) and (d) patch length \(p\) in \sysname for long-term forecasting on PEMS08 dataset.}
  \label{fig:param_study}
\end{figure}

We systematically examine the influence of key hyperparameters on the performance of \sysname, focusing on the input length, embedding size, and the number of prototypes. These parameters are chosen due to their adjustable nature and their substantial impact on the model's architecture and performance.
The analysis is conducted on the PEMS08 dataset, with results visualized in ~\figref{fig:param_study}. 
This evaluation provides insights into the sensitivity of \sysname to these critical hyperparameters, enabling a deeper understanding of its performance and guiding parameter tuning for practical use.

\subsubsection{Impact of Number of Prototypes \(k\)}

\figref{fig:param_study_k} depicts the impact of varying the number of prototypes \(k\) in \sysname on the PEMS08 dataset. 
Increasing \(k\) enlarges the prototype correlation matrix we need to calculate in the online phase, driving a predictable and steady increase in both FLOPs and peak memory consumption. 
At the same time, more prototypes can capture more meaningful correlations, enhancing model performance to a degree.
Notably, once \(k\) exceeds a certain threshold, further gains in prediction accuracy plateau, aligning with our motivation in \secref{sec:motivation}, which highlights that the segment patterns in the data are not overly complex. 

\subsubsection{Impact of Embedding Size \(d\)}

\figref{fig:param_study_d} illustrates how the embedding size \(d\) in \sysname affects performance on the PEMS08 dataset. 
As \(d\) increases, both FLOPs and peak memory usage rises while prediction accuracy gradually improves.
Specifically, as \(d\) grows, we observe a steady decline and eventual convergence of prediction error.
However, this improvement comes at a steep cost: the marginal gains in accuracy diminish rapidly as the computational overhead continues to escalate. 
This insight suggests that the selection of  an appropriate embedding size should be guided by a balance between performance needs and resource constraints, tailored to the specific application context.

\subsubsection{Impact of Input Window Size \(L\)}

\figref{fig:param_study_l} shows how the input window size \(L\) influences performance in \sysname on the PEMS08 dataset. 
Extending the input window increases the information available to the model, consistently leading to a reduction in prediction error. 
However, unlike the trend observed with embedding size scaling, we see a stable improvement in predictive capability. 
This enhancement comes at a cost: processing longer input sequences inflates both FLOPs and peak memory requirements. 
This behavior supports a potential scaling law in time series prediction, as suggested in \cite{shi2024scaling}, which posits that longer input sequences can yield substantial performance gains. 
\sysname adheres well to this principle, demonstrating its strong potential for scaling effectively to exploit ultra long input sequences.

\subsubsection{\blue{Impact of Patch Length \(p\)}}

\blue{
\figref{fig:param_study_p} shows the impact of patch length on model accuracy and performance. Shorter patches boost performance by generalizing to diverse temporal patterns as we mentioned in \secref{sec:motivation}, but increase processing overhead as more patches are needed for the prediction.
This finding underscores the importance of striking a balance between the model's generalization ability and efficiency when selecting the patch length.
}

\subsection{Ablation Study} 
\label{sub:ablation_study}
We conducted an ablation study on the PEMS04 dataset to quantify the impact of each component in \sysname, assessing both accuracy and efficiency through several model variants: 
\begin{itemize} 
\item \textit{\sysname-Attn}:The feature extractors are replaced with Self-Attention layers. 
\item \textit{\sysname-LnrFusion}: The Parallel Fusion Module is replaced by a gated Linear layer.
\item \textit{\sysname-AllLnr}: Both feature extractors and The Parallel Fusion Modules are replaced by Linear layers.
\end{itemize}

As shown in \tabref{tab:ablation_study}, substituting the online modeling process in the extractors with Self-Attention (\sysname-Attn) increases complexity, resulting in higher FLOPs and peak memory usage, but provides negligible improvements in MSE and MAE.
Replacing the Parallel Fusion Module with gated linear layers (\sysname-LnrFusion) reduces accuracy, highlighting the effectiveness of our feature fusion strategy.
The fully linear variant (\sysname-AllLnr), while the most efficient in terms of FLOPs and peak memory usage, shows the poorest accuracy. A comparison between \sysname-AllLnr and \sysname-LnrFusion reveals that the online modeling process introduces only minimal additional resource demands than linear layer.
These findings demonstrate that our design achieves an effective balance between performance and resource efficiency.

\begin{table}[t]
  \centering
  \caption{Ablation Study}
  \label{tab:ablation_study}
  \scriptsize
  \setlength{\tabcolsep}{2pt} 
  \begin{tabular}{c|c|cc|ccc}
    \hline
    \textbf{Dataset} & \textbf{Model} & \textbf{MSE} & \textbf{MAE} & \textbf{FLOPs(M)} & \textbf{Mem(MB)} & \textbf{Param(K)}  \\
    \hline
    \hline
    \multirow{4}{*}{PEMS08}
        &\sysname           & \textbf{0.0711} & \textbf{0.168} & 673 & 79.23 & \underline{1227} \\
        &\sysname-Attn      & \underline{0.0864} & \underline{0.178} & 1235 & 96.13 & 1233\\
        &\sysname-LnrFusion & 0.0875 & 0.191 & \underline{559} & \underline{54.93} & 1438 \\
        &\sysname-AllLnr    & 0.0897 & 0.193 & \textbf{519} & \textbf{50.96} & 1429 \\
    \hline
    \multirow{4}{*}{Electricity}
        &\sysname           & \textbf{0.162} & \textbf{0.258} & 3929 & 266 & \underline{2617} \\
        &\sysname-Attn      & \underline{0.163} & \underline{0.258} & 4434 & 305 & 2650 \\
        &\sysname-LnrFusion & 0.167 & 0.264 & \underline{3134} & \underline{177} & 2989 \\
        &\sysname-AllLnr    & 0.173 & 0.281 & \textbf{2966} & \textbf{168} & 2956 \\
    \hline
\end{tabular}
\end{table}

\begin{figure}[t]
  \centering
\subfloat[PEMS08]
  {
      \label{fig:correlation_loss_1}\includegraphics[width=0.42\linewidth]{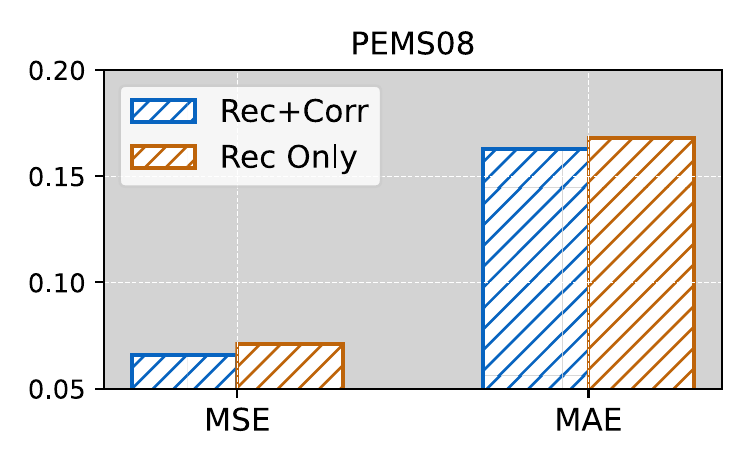}
    }
  \subfloat[Electricity]
  {
      \label{fig:correlation_loss_2}\includegraphics[width=0.42\linewidth]{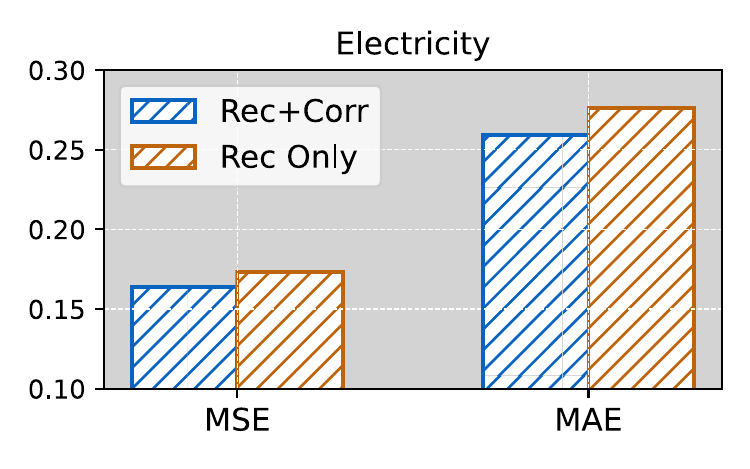}
  }
  \caption{Comparison of accuracy between prototypes optimized only based on Euclidean distance reconstruction error (\textit{Rec Only}) and those incorporating correlation error  (\textit{Rec+Corr}).}
  \label{fig:offline_study}
\end{figure}

\blue{
\subsection{Study towards generalization on test set}
\label{sub:generalization_test}
The non-stationary property of time series data often causes new segment patterns to emerge in the test set. 
This challenges the model's generalization ability.

To test its impact on model performance, we used the t-SNE \cite{van2008visualizing} method to compare segment distributions in the training and test sets of the Electricity dataset. 
We then identified test-set instances containing unseen segments to test the model's forecasting accuracy, comparing with PatchTST (also segmentation-based), as shown in \figref{fig:shift}.
The results indicate that the input sequences contain unseen segments, mainly characterized by steeper intra-segment trends.
FOCUS predicts larger data-change magnitudes than PatchTST, better following ground truth trends.
In terms of the method, because the clustering process of FOCUS assists the model in associating new segments with known segments, it can reduce the difficulty for FOCUS to understand new segment patterns.

\begin{figure}[t]
  \centering
  \includegraphics[width=\linewidth]{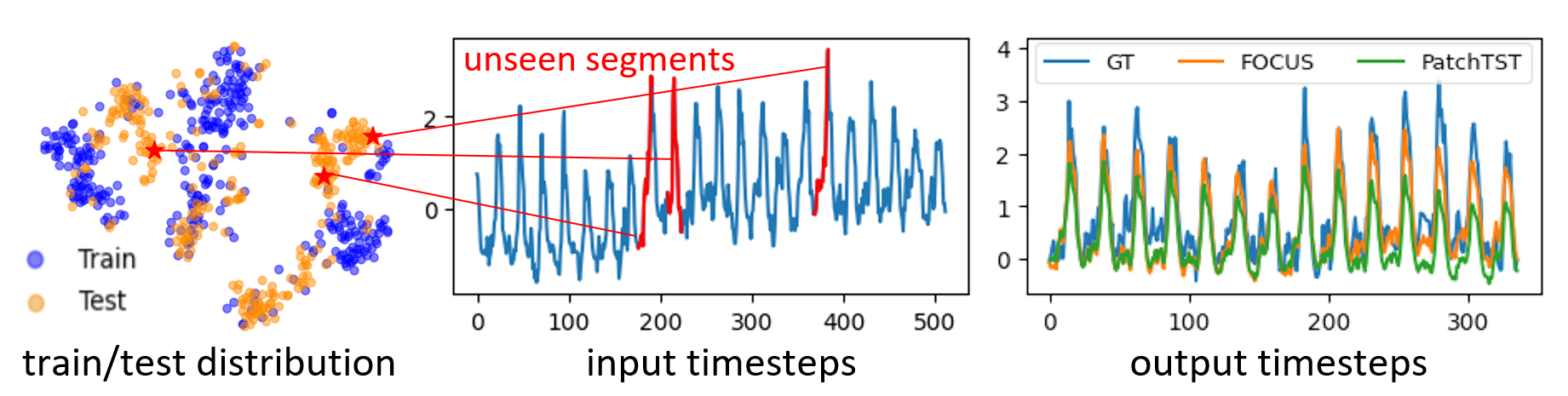}
  \caption{A test instance in the Electricity dataset and the prediction results of FOCUS and PatchTST on it.}
  \label{fig:shift}
\end{figure}
}

\blue{
\subsection{Study towards the effect of outliers}
\label{sub:robustness_test}
When collecting time series data, outliers often occur due to collection device reliability issues. 
These outliers can harm the model's modeling ability, so it's crucial to test the model's outlier resistance.
We simulate different ratio by replacing training data points with outliers (sampled from a distribution over three-times the real data's standard deviation), as shown in \figref{fig:perturb_example} and then test the model's forecasting accuracy.

As shown in \figref{fig:noisy_ratio_comparison}, the prediction accuracy of \sysname stays relatively stable under a certain degree of perturbation. 
There is a notable increase when the perturbation ratio is around 10\%. 
A similar phenomenon is observed in PatchTST, where its accuracy spikes when the perturbation ratio is about 6\%.
By comparison, \sysname has stronger ability to resist outliers. 
In terms of the method, FOCUS assigns data segments to the nearest clustering centers.
This operation helps to minimize the impact of outliers, ensuring the performance remains stable even as the perturbation ratio improves.

\begin{figure}[t]
  \centering
  \subfloat[Perturbation]
  {
    \label{fig:perturb_example}\includegraphics[width=0.32\linewidth]{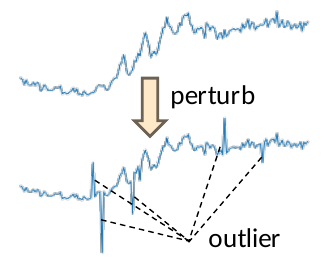}
    }
  \subfloat[Performance degradation]
  {
  \label{fig:noisy_ratio_comparison}\includegraphics[width=0.55\linewidth]{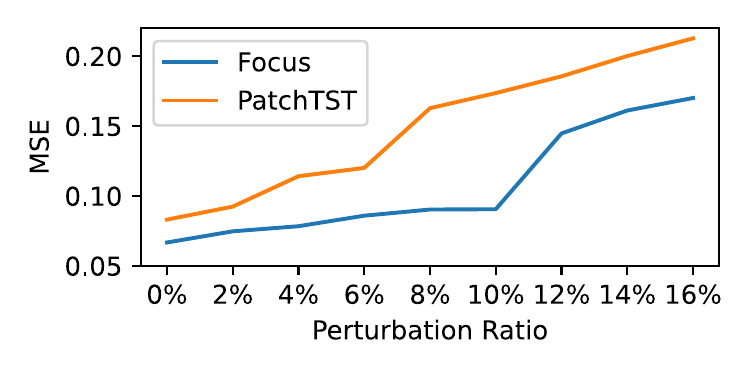}
    }
  \caption{The forecasting accuracy under different perturbation ratios.}
  \label{fig:noise_test}
\end{figure}
}



\subsection{Experiment on Offline Processing}
\label{sub:offline_processing}
In the offline clustering phase, we utilize reconstruction error based on Euclidean distance and correlation error based on Pearson correlation to identify and optimize typical segment patterns in the dataset, generating prototypes for use in the online phase. 
To evaluate the impact of clustering objectives, we compare two approaches for obtaining prototypes: clustering optimized solely with reconstruction error (Rec Only) and clustering optimized with both reconstruction and correlation errors (Rec+Corr) on PEMS08 and Electricity datasets, while keeping all other settings consistent.
Since the primary goal of the prototypes is to enhance prediction accuracy during the online phase, we use the prediction accuracy of the final model trained with each set of prototypes as the evaluation metric.

\figref{fig:offline_study} illustrates the effects of correlation loss. 
Models utilizing prototypes obtained with correlation error demonstrate improved prediction accuracy in terms of both MSE and MAE.
\blue{Meanwhile, we observe that the additional running time is indistinguishable from noise, which means that for the proposed method, the improvement is almost cost-free.}
This indicates that incorporating correlation error during the offline phase yields prototypes that better support long-sequence modeling and prediction in the online phase.

\begin{figure}[t]
  \centering
  \includegraphics[width=0.8\linewidth]{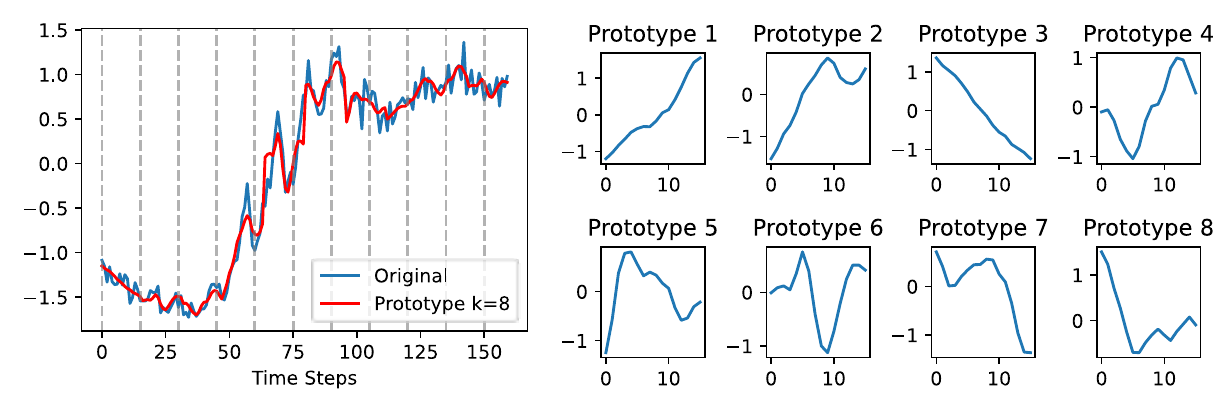}
  \caption{Series approximation of the rise at morning with the number of prototypes \(k=8\).}
  \label{fig:visualization_prototypes}
\end{figure}

\subsection{Case Study}
\label{sub:case_study}

\begin{figure}[t]
  \centering
  \subfloat[Input Series]
  {
      \label{fig:case_study_inputs}\includegraphics[width=0.5\linewidth]{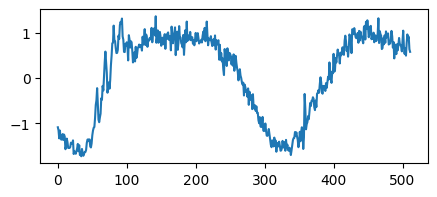}
    }
    \hfill
  \subfloat[Forecasting Result]
  {
      \label{fig:case_study_prediction}\includegraphics[width=0.4\linewidth]{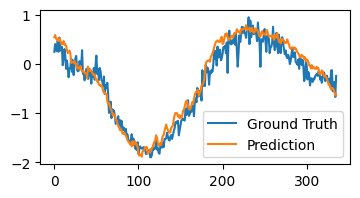}
    }
  \caption{Visualization of the input series for case study and corresponding forecasting result.}
  \label{fig:case_study_forecasting}
\end{figure}

\begin{figure}[t]
  \centering
  \includegraphics[width=0.7\linewidth]{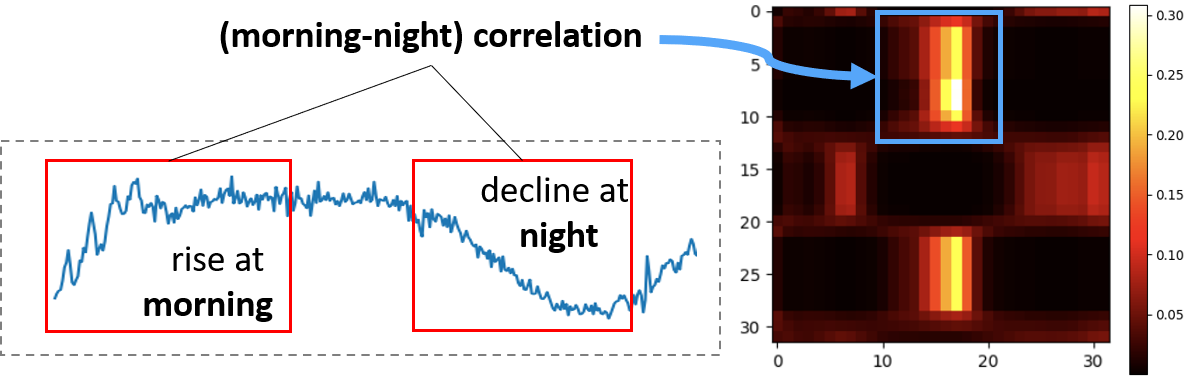}
  \caption{An example of long-range dependency extracted by \sysname.}
  \label{fig:case_study_correlations}
\end{figure}

We conducted an in-depth case study on a randomly sampled sequence from the PEMS08 dataset, which is shown in ~\figref{fig:case_study_inputs},  evaluating the quality of the model's predictions, how the model interprets the sequence dynamics, and analyzing how prototypes effectively approximate long sequences.

\fakeparagraph{Forecasting Result}
\figref{fig:case_study_prediction} presents the model's forecasting results. 
The predictions closely match the ground truth, even for subtle patterns, such as the slight rise before the traffic decline and multiple spikes during the morning rise. The ability to predict these intricate details indicates the model’s proficiency in understanding long-term dependencies and the sequence's nuanced behaviors.

\fakeparagraph{Learned Long-range Dependency}
We further examined the temporal dependencies learned by \sysname, obtained by directly multiplying the assignment matrix with the online correlation matrix, as shown in ~\figref{fig:case_study_correlations}.
Notably, the analysis highlights a strong dependency between the rise of traffic flow in the morning and the decline in the night. 
This insight confirms that our model captures significant long-range dependencies and understands the underlying temporal dynamics of traffic patterns.

\fakeparagraph{Approximation via Prototypes}
To approximate the original time series effectively, we decomposed the sequence into $k=8$ prototypes, with each prototype adjusted to maintain the original mean and standard deviation. 
As illustrated in \figref{fig:visualization_prototypes}, these prototypes capture critical features of the sequence, such as the distinct spikes observed between the 16th and 64th time steps. This demonstrates that a limited set of prototypes, when combined with local statistical details, can capture and approximate the essential patterns of complex time series data.



\subsection{Summary of Major Experimental Findings}
The major experimental findings are summarized as follows:
\begin{itemize}
\item \sysname achieves outstanding accuracy with minimal FLOPs compared to a diverse and comprehensive set of baselines across multiple datasets, while still maintaining lower peak memory usage.
\item Parameter study reveal that \sysname maintains feasible computational overhead and consistently improves prediction accuracy when scaling to longer time series.
\item The case study highlights the effectiveness of \sysname's approximation for long sequence representation and its ability to capture meaningful long-range dependencies.
\end{itemize}

\section{Related Work}
\label{sec:related}

\fakeparagraph{Multivariate Time Series Forecasting}
Multivariate Time Series (MTS) forecasting, essential for modeling the temporal dynamics of multiple variables, has garnered significant attention~\cite{qiu2024tfb} due to its applications in various domains ~\cite{deng2021st, cirstea2022towardstraffic, cirstea2021enhancenet, hu2020stochastic, cheng2021windturbine, lai2018energy,wang2019weatherforecasting, he2022climateforecasting, chen2022reliable, chen2024playbest}.

Statistical models like ARIMA~\cite{ariyo2014arima} and VAR~\cite{kilian2017var} assume linear dependencies, limiting their effectiveness for complex, high-dimensional data.
Deep learning approaches, including LSTNet~\cite{li2020lst} and Wavenet~\cite{van2016wavenet}, leverage CNNs or RNNs to model nonlinear spatial and temporal patterns. However, these methods struggle with long-term dependencies and scalability.Spatial-Temporal Graph Neural Networks (STGNNs) like AGCRN~\cite{bai2020agcrn}, METRO~\cite{cui2021metro}, Graph Wavenet~\cite{wu2019graphwavenet}, and MTGNN~\cite{wu2020mtgnn} integrate graph convolutions for spatial dependencies. 
While effective, their limited receptive fields hinder long-term forecasting.

Transformers overcome these limitations with self-attention mechanisms, excelling at modeling long-term dependencies. Informer~\cite{zhou2021informer} improves efficiency with ProbSparse attention, Autoformer~\cite{wu2021autoformer} enhances interpretability via decomposition, and PatchTST~\cite{nie2023patchtst} scales to high dimensions with patch-based representations. Crossformer~\cite{zhang2023crossformer} integrates temporal and inter-variable attention, setting new benchmarks.

\blue{
\fakeparagraph{Efficient Long-Range Dependency Modeling}
Handling long-range dependencies is a fundamental challenge in multivariate time series, as these dependencies capture essential relationships across time and variables. 
Transformers revolutionized sequence modeling by capturing long-range dependencies. 
However, their quadratic complexity ($O(L^2)$) limits the scalability for time series data \cite{zhou2021informer}.

Recent work aims to optimize processing performance with different inductive biases. Informer \cite{zhou2021informer} prunes low-value info using the attention mechanism's low-rank property, which cuts costs but risks losing information. 
Pyraformer \cite{liu2021pyraformer} tries to lower complexity in a hierarchical way, though this introduces propagation errors.
FedFormer \cite{zhou2022fedformer} uses time-series data's frequency-domain features, but frequency-domain sampling may lose important info.
PatchTST \cite{nie2023patchtst} and Crossformer \cite{zhang2023crossformer} discretize time-series data by patching. This improves prediction efficiency and reaches top-notch accuracy. However, patching can't reduce the time-series dimension complexity.

Based on discretization, \sysname starts from the time series data itself. It attains the discretized representation by clustering time-series segments based on their similarity, enables time-series modeling and prediction with linear complexity.
}

\section{Conclusion}
\label{sec:conclusion}
In this paper, we introduce \sysname, a novel two-phase multivariate time series forecasting model. 
It aims to balance accuracy and computational efficiency when modeling long-range dependencies.
\sysname combines an offline clustering phase with an online adaptation phase. The offline phase identifies representative segment patterns from the whole dataset, capturing key time-series features. These patterns are then dynamically adjusted in the online phase to adapt to trends and capture long-range dependencies in the input data.
This integration enables \sysname to efficiently capture complex long-range dependencies while reducing computational overhead. Experiments on various datasets show that \sysname has significant advantages in forecasting accuracy and computational efficiency compared to SOTA models. Its robustness and adaptability make it an attractive choice for practical applications, especially in resource-constrained environments.

\section{Acknowledgements}

This work was partially supported by National Key Research and Development Program of China under Grant No. 2023YFF0725103, National Science Foundation of China (NSFC) (Grant Nos. 62425202, U21A20516, 62336003), the Beijing Natural Science Foundation (Z230001), the Fundamental Research Funds for the Central Universities No. JK2024-03, the Didi Collaborative Research Program and the State Key Laboratory of Complex \& Critical Software Environment (SKLCCSE).
Zimu Zhou's research is supported by Chow Sang Sang Group Research Fund No. 9229139.
Jinliang Deng's research is supported by Theme-based Research Scheme (T45-205/21-N) from Hong Kong RGC, Generative AI Research and Development Centre from InnoHK, and the Open Project Program of State Key Laboratory of Virtual Reality Technology and Systems, Beihang University (No.VRLAB2024A02).
Corresponding authors are Yongxin Tong and Jinliang Deng.

\clearpage
\bibliographystyle{IEEEtran}
\balance
\bibliography{ref}

\end{document}